\documentclass{article} 
\usepackage[final]{colm2025_conference}
\usepackage{microtype}
\usepackage{hyperref}
\usepackage{url}
\usepackage{booktabs}
\usepackage{graphicx}
\usepackage{lineno}

\usepackage{array,tabularx}
\usepackage{multirow}
\newcolumntype{L}{>{\raggedright\arraybackslash}X} 

\definecolor{darkblue}{rgb}{0, 0, 0.5}
\hypersetup{colorlinks=true, citecolor=darkblue, linkcolor=darkblue, urlcolor=darkblue}

\title{Mic Drop or Data Flop? Evaluating the Fitness for Purpose of AI Voice Interviewers for Data Collection within Quantitative \& Qualitative Research Contexts
}

\author{Shreyas Tirumala, Nishant Jain \& Danny D. Leybzon \\
VKL Research, Inc.\\
San Francisco, CA \\
\texttt{\{shreyas,nishant,danny\}@vkl.ai} \\
\AND
Trent D. Buskirk \\
Professor, Provost Fellow of Data Science \\
Old Dominion University \\
Norfolk, Virginia \\
\texttt{tbuskirk@odu.edu} \\
}

\begin{document}

\ifcolmsubmission
\linenumbers
\fi

\maketitle

\begin{abstract}
Transformer-based Large Language Models (LLMs) have paved the way for "AI interviewers" that can administer voice-based surveys with respondents in real-time. This position paper reviews emerging evidence to understand when such AI interviewing systems are fit for purpose for collecting data within quantitative and qualitative research contexts. We evaluate the capabilities of AI interviewers as well as current Interactive Voice Response (IVR) systems across two dimensions: input/output performance (i.e., speech recognition, answer recording, emotion handling) and verbal reasoning (i.e., ability to probe, clarify, and handle branching logic). Field studies suggest that AI interviewers already exceed IVR capabilities for both quantitative and qualitative data collection, but real-time transcription error rates, limited emotion detection abilities, and uneven follow-up quality indicate that the utility, use and adoption of current AI interviewer technology may be context-dependent for qualitative data collection efforts.
\end{abstract}

\section{Introduction}

\subsection{Background}

Since \citet{Vaswani+2017}'s seminal work on transformers revolutionized natural language processing, Large Language Models (LLMs) have become increasingly commonplace tools to parse and produce human language. As such technologies have matured, survey methodologists, too, have begun to explore their potential \citep{lerner2024, buskirk2025}. One area of inquiry, in particular, involves understanding the feasibility of using artificial intelligence (AI) interviewers that leverage LLMs for data collection.

Automated systems have long been studied as a means of improving data quality while reducing costs. Years of social desirability bias research suggest that, particularly for sensitive topics, respondents may be more forthcoming in automated, self-administered surveys than in traditional interviews with other humans \citep{gribble1999interview, cooley2000automating, kreuter2008social, goldstein2025digging}. The open question, however, is what logic such automated systems should use to elicit the highest quality responses.

Although early automated systems were more akin to rules engines that followed pre-defined logic to decide how to respond, more recent work has explored more expressive "chatbots" that use LLMs to communicate with respondents dynamically using a text-based chat interface. \citet{kim2019comparing} show that employing such technologies resulted in lower rates of "satisficing" (selecting the easiest possible valid option) among respondents. Later work by \citet{rhim2022application}, \citet{chopra2023conducting}, and \citet{barari2025ai} support these findings, with Rhim and colleagues observing that more conversational chatbots yielded higher levels of respondent self-disclosure – at the expense of slightly elevated social desirability bias.

In recent years, however, voice interfaces have received newfound attention. Even before the advent of transformer-based LLMs, audio interviews had been explored as a means of increasing self-disclosure \citep{lind2013survey, lucas2014s, devault2014simsensei}. Voice is a particularly compelling modality because some respondents may be more forthcoming via voice than via web forms \citep{galasso2024we}. Additionally, audio-based interviews allow researchers to record an entire survey interview as it is conducted, which can be a useful strategy for quality control. \citet{gomila2017audio} suggest that collecting and analyzing recordings helps mitigate response fraud, a technique unavailable for text-based surveys.

Voice interviews are not a panacea, however. Mode effects may affect the types of responses collected. For instance, respondents may be more likely to round numerical answers via voice compared to text-based input methods \citep{schober2015precision}. Moreover, from an operational perspective, some types of questions are easier to communicate textually than orally; for example, multi-select questions (i.e., "Select all that apply") often involve large numbers of answer choices that are cumbersome to be heard than read. Such questions also pose data quality concerns. Some researchers have questioned whether respondents tend to over-index on the first or last answer choices presented in voice surveys (i.e., primacy or recency effects), especially as the number of choices provided increases \citep{le2013investigating}.

Recent work by \citet{leybzon2025ai}, \citet{lang2025telephone}, and \citet{wuttke2024ai} establishes that AI interviewers powered by LLMs can indeed conduct voice-based interviews. This begs a natural question: When and how should contemporary AI voice interviewers be used? In this work, we aim to address this question and examine the fitness for purpose of contemporary voice AI technologies for collecting data within both quantitative and qualitative research contexts.

\subsection{Definitions}

While the technical language we enumerate here may be familiar to computer scientists and survey researchers, we provide a selected list of terms and definitions to facilitate a common language for analysis.   

\subsubsection{Interactive Voice Response (IVR)} 

IVR is a technology that uses a set of pre-recorded audio messages to communicate with a user over the phone \citep{corkrey2002interactive}. Most IVR systems rely on touch-tone inputs (i.e., numbers dialed on a keypad) to record user selections and responses. Based on a user's inputs, the system selects its next response using pre-defined logic (i.e., a decision tree or flowchart). IVR is commonly used in both customer service and telephone survey contexts.

\subsubsection{AI Interviewer}
In this paper, we will use the term "AI interviewer" to refer to a system that leverages computational techniques to query, hear, understand, respond and interact with survey respondents via audio without relying entirely on preset scripts. This evaluation will focus on nascent audio-based systems rather than text-based chatbots.

Advanced contemporary systems typically comprise the following primary components as described here and illustrated in Figure 1:
\begin{samepage}
\begin{itemize}
    \item \textbf{Real-time Automated Speech Recognition (ASR):} Sometimes referred to as Speech-to-Text (STT), these systems transcribe respondent audio. ASR provides the "ears" of an AI interviewer, enabling them to hear respondents.
    \item \textbf{Large Language Models (LLMs):} Transformer-based LLMs are the "brains" of an AI interviewer; they interpret respondent input in order to make decisions mid-conversation and decide what to say back.
    \item \textbf{Speech Synthesis:} Also called "Text-to-Speech" (TTS), these technologies serve as the "mouth" of an interviewer system by producing human-sounding voices to speak back to a respondent based on text sent to it by an LLM.
\end{itemize}
\end{samepage}
\begin{figure}[ht]
\begin{center}
\includegraphics[width=0.8\linewidth]{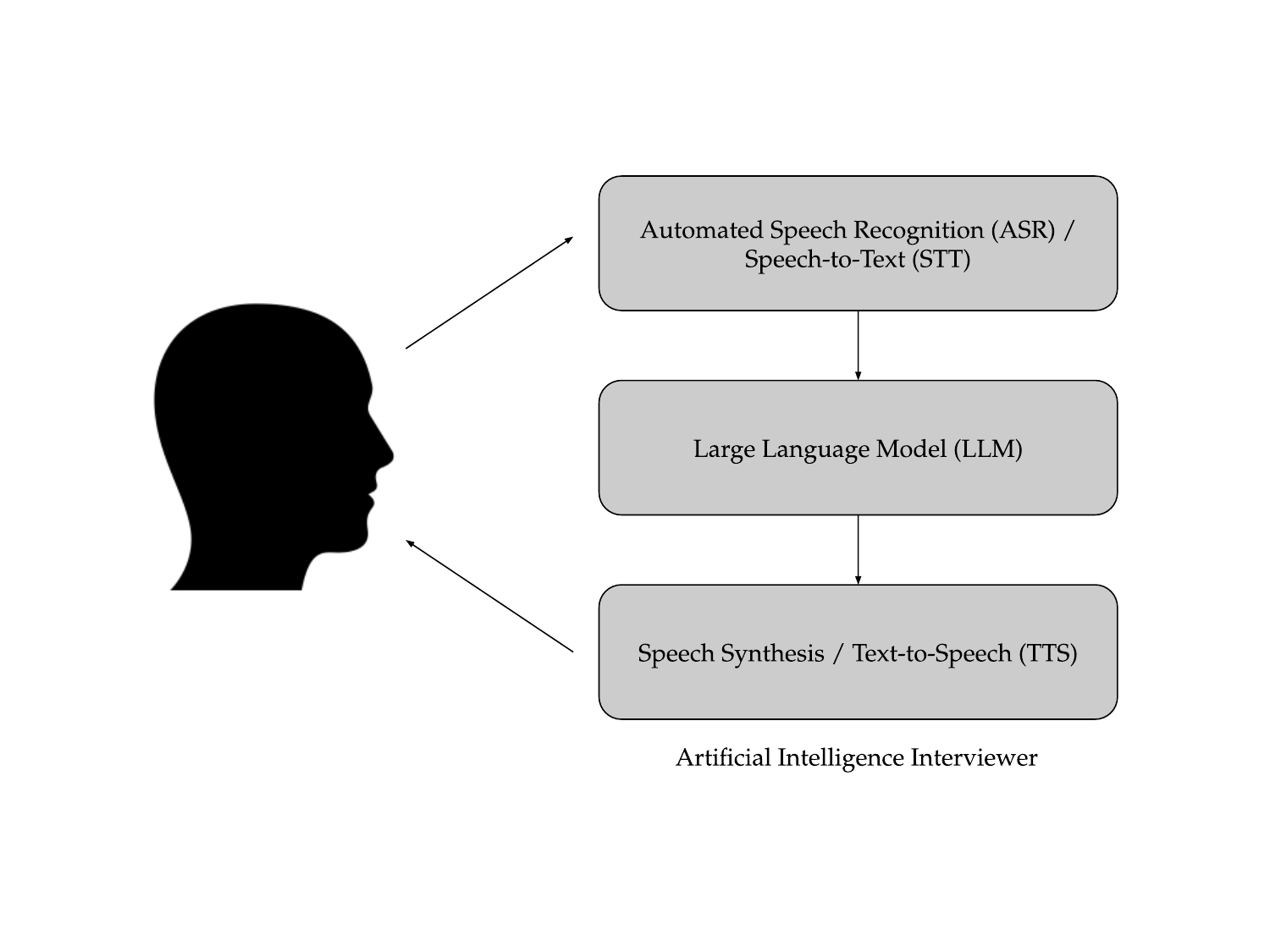}
\end{center}
\caption{Standard AI interviewer architecture}
\end{figure}

As depicted in Figure 1, by connecting ASR systems to LLMs and speech synthesis technologies, a conversational AI agent can replicate a human's ability to ask questions, hear answers, and create appropriate responses. Although additional software infrastructure components (e.g., text output safeguards such as profanity filters, phone call deployment automation) are typically used to connect AI interviewer components together, we consider such technologies to be outside the scope of this analysis, as they do not meaningfully impact an AI interviewer's core capabilities.

Though preliminary work has explored the usage of models to generate output audio directly without an intermediate conversion to text (see \citet{chu2023qwenaudioadvancinguniversalaudio, tang2024salmonngenerichearingabilities, xu2025qwen25omnitechnicalreport}), the feasibility of such speech-to-speech models in AI interviewer systems has not yet been successfully demonstrated. To date, such models introduce too much latency for real-time conversational contexts \citep{cui2025recentadvancesspeechlanguage}, though architectural improvements have been proposed in research settings to address this limitation \citep{défossez2024moshispeechtextfoundationmodel}.

\subsubsection{Quantitative Data Collection}
Quantitative surveys typically involve a pre-defined list of closed-ended questions and pre-set lists of valid answer choices/criteria. Quantitative surveys often use multiple choice questions, rating scale questions (e.g., Likert), numerical questions, and binary yes/no questions, but may also request some limited open-ended responses. Respondent data is typically aggregated and analyzed for quantitative results rather than qualitative themes \citep{groves2011survey}.
  
\subsubsection{Qualitative Data Collection}
Qualitative data collection takes many forms, including in-depth interviews, focus groups, cognitive pre-testing, and more. Many methods of qualitative data collection also leverage a pre-defined list of questions or topics; however, they are more likely to feature longer open-ended questions that invite respondents to expand upon their beliefs, as well as unscripted follow-ups \citep{adams2015conducting}. Such methods have historically leveraged human judgment (e.g., emotional state evaluations of respondents, semantic evaluations of how vague or ambiguous answers are) to determine which topics require additional investigation mid-conversation. The importance of human intervention has made such conversations difficult to automate.

\section{Evaluation Criteria}

We will assess the fitness for purpose of IVR and AI interviewer technologies for data collection within both quantitative and qualitative research contexts. Specifically, using qualitative evaluations of fit, we examine whether the data collected by each system meets standard research objectives across two quality domains: Input/Output and Verbal Reasoning \citep{bosch2023utility}. Table 1 lists key questions we evaluate.

\begin{table}[ht]
\centering
\renewcommand{\arraystretch}{1.25}

\begin{tabularx}{\linewidth}{|L|L|L|}
\hline
\multirow{2}{=}{\textbf{Data Collection Context}}  
 & \multicolumn{2}{c|}{\textbf{Quality Domain}} \\ \cline{2-3}
 & \textbf{Input/Output:} \textit{Can a system adequately receive answers and reply back to respondents?} &
   \textbf{Verbal Reasoning:} \textit{How intelligently can a system alter the conversation based on previous user input?} \\ \hline
\textbf{Quantitative Data Collection:} \textit{Surveys with pre-set lists of questions and answers; these typically involve closed-ended questions} &
Are questions asked and responses given faithful to the original survey design?\par\bigskip
Are answers collected faithful to a respondent's intent? &
What types of conversational choices can each system make based on a respondent's answers (e.g., branching logic, clarifications, corrections)?\par\bigskip
How do the conversational choices that each system can make impact data quality? \\ \hline
\textbf{Qualitative Data Collection:} \textit{Semi-structured interviews with several open-ended questions and unscripted follow-up questions} &
How well can each system handle open-ended questions?\par\bigskip
Can each system identify emotional context and vary its output accordingly? &
Can each system probe — that is, produce unscripted follow-up questions?\par\bigskip
If so, are these follow-up questions useful for improving response quality and depth? \\ \hline
\end{tabularx}

\caption{Comparison of input/output and verbal reasoning domains across data collection contexts}
\label{tab:io-vr}
\end{table}

\section{Capability Evaluations}

In the following section, we will assess both IVR and AI interviewer systems. For each of the quality domains listed in Table 1, we will begin by providing an overview of each system's capabilities followed by discussions of its fitness for purpose for both quantitative and qualitative research.

\subsection{Interactive Voice Response (IVR)}
To understand the functionality of more advanced AI interviewing systems, we must first understand the well-established technologies they are intended to replace. Automated phone-based surveys have been traditionally handled by Interactive Voice Response (IVR) systems.

\subsubsection{Input/Output}
\textbf{Capability Overview:} The most commonly used systems enable respondents to select answer choices by dialing numbers on a phone keypad via a touch-tone interface. These systems play back pre-recorded audio samples based on respondent selections. More recent systems supplement the touch-tone interface with ASR and leverage text-to-speech to varying degrees of success, though such systems are less common \citep{inam2017comparative}. In many cases, a recorded message must play back in its entirety before a response will be accepted.

\textbf{Fitness for Purpose (Quantitative Data Collection):} The touch-tone interface is straightforward to use for respondents, but poses an issue for survey researchers. Due to limitations of touch-tone interfaces, quantitative survey questionnaires must often be adapted for use with IVR systems. Question wordings are often changed to add additional instructions; for example, to ask about a respondent's gender identity, rather than asking "Do you currently identify as a man, a woman, or in some other way?", \citet{amaya2021call} asked the following question instead: "Do you currently identify as a man? Press one. For a woman, press two, or for some other way, press three."  Although such changes may be acceptable in some contexts, doing so may complicate comparative analysis between modes in mixed-mode projects.

\textbf{Fitness for Purpose (Qualitative Data Collection):} Structural challenges prevent effective qualitative data collection via IVR. Asking open-ended questions, for instance, requires researchers to perform post-hoc transcriptions of audio recordings of each open-ended response \citep{amaya2021call}. Focus groups and multi-party interviews are also not feasible with IVR systems as touch-tone interfaces are designed for one individual to use at a time. Given the pre-recorded nature of IVR systems, it is not possible to vary output back to a respondent based on emotional cues, though post-hoc analysis of voice recordings may be used to identify these after the survey.

\subsubsection{Verbal Reasoning}
\textbf{Capability Overview:} IVR systems typically have limited verbal reasoning capabilities; mid-conversation decision-making must be pre-defined with any desired follow-ups or responses recorded in advance \citep{kim2011inherent}.

\textbf{Fitness for Purpose (Quantitative Data Collection):} The lack of dynamic response capabilities in IVR systems inhibits their ability to conduct quantitative surveys. Respondents cannot usually clarify answer choices, correct previous responses, or repeat specific portions of a question \citep{kim2011inherent}. Unlike trained human interviewers, IVR systems also do not typically offer respondents encouragement mid-survey to facilitate completion \citep{amaya2021call}.

These limitations may impact data quality and completion rates. The inability to clarify or repeat portions of a question opens the door to order effects. Particularly for lengthy questions, respondents may forget answer choices, creating a bias toward the first or last few options spoken. Even after controlling for such order effects, \citet{dillman2009response} find that IVR respondents tend to give more extreme positive responses than other modes, and \citet{amaya2021call} notes higher rates of both misreporting and item non-response.

\textbf{Fitness for Purpose (Qualitative Data Collection):}
Because IVR systems lack the ability to dynamically react to a respondent's speech input, it is not possible to conduct any interviews that necessitate contextually-determined follow-up questions. Such functionality is crucial for in-depth interviews and cognitive pretesting. Accordingly, IVR systems have little utility for this specific type of qualitative data collection.

\subsection{AI Voice Interviewers}
Contemporary AI interviewers address many, but not all, of the issues raised by IVR. The verbal reasoning capabilities that LLMs provide to these systems enable follow-up questions and complex decisionmaking. In the last year, \citet{leybzon2025ai}, \citet{lang2025telephone}, and \citet{wuttke2024ai} have all been able to successfully and independently conduct quantitative voice surveys using AI interviewers. These works serve as examples of AI interviewer capabilities at this time and form the core of our analysis. 

Because research in this field remains nascent and many questions are still unanswered, we augment our assessment of AI interviewers with insights drawn from the broader machine learning literature.

\subsubsection{Input/Output}
\textbf{Capability Overview:} Unlike the touch-tone interface of IVR systems, AI interviewers use ASR and speech synthesis to collect respondent input and vocalize responses. 

Contemporary ASR systems have achieved relatively high-levels of accuracy. \citet{kuhn2024measuring} find that for state-of-the-art ASR systems, English word error rates hover around 5\%, though heavily accented speech may decrease performance. Further research is needed to understand how error rates vary for other languages.

On the speech synthesis side, modern systems are advanced enough to reproduce accented English speech \citep{zhou2024accented}. Many speech synthesis systems allow for AI interviewers to select voices on a per call basis. This opens the possibility for intelligent voice selection based on call-specific factors (e.g., individual respondent demographics). Though such experiments have not been conducted to date, work by \citet{lilley2025social} suggests that doing so may produce similar effects to varying which human interviewers administer surveys.

The accuracy of ASR and the sophistication of speech synthesis systems allow AI interviewers to react intelligently to respondents using voices that sound relatively human. Such systems have been shown to be robust enough to handle respondent interruptions and pauses mid-survey \citep{leybzon2025ai}.

Two notable blind-spots for such systems, however, are emotion detection and expression. Because AI interviewers to-date have converted audio to text via ASR, the tone of the speaker must be inferred based on text. Even if emotion detection were to be implemented in such systems, \citet{alhussein2025speech} find that existing AI-based emotion detection systems yield mixed results. Vocal expression of emotion is also difficult for modern speech synthesis systems. Though theoretical frameworks have been proposed to enable artificial voices to vary both emphasis and prosody, consumer-grade speech synthesis tools implementing these frameworks have not yet been studied in a survey context \citep{seshadri2021emphasis}.

\textbf{Fitness for Purpose (Quantitative Data Collection):} The findings of \citet{leybzon2025ai}, \citet{lang2025telephone}, and \citet{wuttke2024ai} suggest that word error rates for modern ASR systems are at least sufficient to conduct quantitative surveys, however some caveats do exist. 

\citet{kuhn2024measuring} observe that although word error rates are low for state-of-the-art systems, real-time/streaming transcription error rates can be significantly higher (\textasciitilde10.9\% on average). Indeed, \citet{wuttke2024ai} and \citet{leybzon2025ai} both note that respondents reacted to minor speech recognition errors mid-survey. However, problems with erroneous transcriptions can be somewhat mitigated by post-processing recorded audio after the survey is over.

Speech synthesis also appears to be adequate for quantitative use cases – no AI interviewer study to date has remarked on specific issues associated with TTS systems, even with their limited ability to express emotion or emphasis. Importantly, in comparison to IVR, an AI interviewer can recite question text as-is, without requiring wording changes to suit the medium; that is, both human and AI interviewers can ask the same set of questions.

\textbf{Fitness for Purpose (Qualitative Data Collection):} High streaming transcription error rates and a lack of emotion detection/expression pose issues for qualitative data collection. Unlike quantitative data collection methods, qualitative research often involves a high proportion of open-ended questions that lack pre-defined lists of valid responses. Without a list of valid responses to check against, transcription errors cannot easily be corrected. This is especially problematic because open-ended questions also typically yield higher average response lengths than closed-ended multiple choice questions; for instance, asking respondents to elaborate on their political beliefs requires more explanation than asking them to classify their political party affiliations. Longer responses increase the expected number of transcription errors per survey. Further complications arise in focus groups and multi-party interviews as ASR performance declines when two or more speakers are audible \citep{addlesee2020comprehensive}.

Emotion detection and expression are also more important in qualitative data collection contexts than they are in quantitative contexts. Building rapport is crucial for maximizing self-disclosure in qualitative interviews, and AI interviewers may need emotion detection and expression skills to do so \citep{sun2021relationship}. A respondent's emotional state also serves as an important input for verbal reasoning; without the emotional context of a response, the quality of follow-up questions generated by an AI interviewer may be lower. It remains to be seen whether AI interviewers can understand a respondent's emotional state enough to vary intonation, emphasis, and follow-up questions effectively.

\subsubsection{Verbal Reasoning} 

\textbf{Capability Overview:} Work by \citet{zeng2023question} suggests that the verbal reasoning capabilities of LLMs were enough to navigate the conditional branching and termination logic of a semi-structured interview. \citet{leybzon2025ai} demonstrate this in the field by using an LLM-powered AI interviewer to administer the SSRS Opinion Panel Omnibus, a 123-question survey that contained a mix of closed-ended questions (multiple-choice, Likert scale, yes/no) and open-ended questions, along with conditional branching/termination logic. This survey took respondents approximately 30 minutes to complete. Both \citet{lang2025telephone} and \citet{wuttke2024ai} also find, independently, that an AI agent was able to successfully navigate shorter survey instruments without major issues. Notably, AI interviewers can respond to requests for clarification or repetition of questions and answer choices, as well as apply branching logic based on the semantic content of open-ended responses. Such features are unavailable in IVR systems.

LLMs also display contextual awareness. When presented with unclear or ambiguous answers, AI interviewers have been shown to re-prompt respondents intelligently. For example, \citet{lang2025telephone} find that AI interviewers could detect and respond to "numeric answer[s] that [were] out of [a valid] range" and \citet{leybzon2025ai} note that AI interviewers can be instructed to re-prompt respondents who answer with just "Liberal" when answer choices are "Somewhat Liberal" or "Very Liberal." Beyond re-prompting, some research has been conducted on the ability of LLMs to generate intelligent follow-up questions with inconclusive assessments of question quality and utility (see \citet{wei2024leveraging}, \citet{geisen2024prompting}, \citet{kuric2024unmoderated}, \citet{cuevas2024collectingqualitativedatascale}, and \citet{chopra2023conducting}). LLMs can also contextually introduce humor, small talk, and affirmation, similar to a human interviewer, though respondent feedback is mixed \citep{yun2023keeping}.

\textbf{Fitness for Purpose (Quantitative Data Collection):} \citet{leybzon2025ai}, \citet{lang2025telephone}, and \citet{wuttke2024ai} all observe that their AI interviewers successfully conducted quantitative surveys with conditional branching logic. Leybzon and colleagues also highlight the AI interviewer's capability to randomize both question order and answer choice order. Although LLMs have been noted to hallucinate (i.e., respond with "plausible, but non-factual content", see \citet{huang2025survey}), to date, none of the three AI voice interviewer studies report any instances of the interviewer inventing questions/answer choices or asking questions out of order. This suggests that AI interviewers perform at least as well as IVR systems with respect to navigating survey instruments.

AI interviewers also have additional verbal reasoning capabilities that may improve data quality. AI interviewers can facilitate clarifications and corrections upon request. Even without a respondent request, it appears that LLMs are capable of basic data integrity validation. As mentioned previously, AI interviewer systems can flag ambiguous responses and answers outside of pre-defined ranges in real-time, allowing them to re-prompt users for valid data.

\textbf{Fitness for Purpose (Qualitative Data Collection):} Qualitative research is often highly interactive; as we have previously highlighted, contextually aware follow-ups are often crucial for collecting high quality data. Though research has shown that LLMs can generate follow-up questions, the quality of these questions is unclear. \citet{cuevas2024collectingqualitativedatascale} find that LLMs tend not to probe enough to identify personalized examples or understand a respondent's specific motivations. \citet{kuric2024unmoderated} corroborate these findings in a user experience research context, observing that OpenAI's GPT-4 model performs worse at eliciting in-depth feedback than a static, pre-curated list of follow-ups created prior to the conversation. Conversely, \citet{geisen2024prompting} and \citet{chopra2023conducting} both suggest that LLM-generated follow-up questions can improve answer depth. Given these varying perspectives, additional investigation may be required to understand follow-up question quality.

LLMs may be able to be tuned to improve follow-up question quality. \citet{wei2024leveraging} show that different system prompts can dramatically impact interviewer behavior and question generation. It is possible that because most ASR systems do not perform emotion detection, improvements to the emotional context fed as input to LLMs could improve performance.

Aside from follow-up questions, further research is necessary to understand whether AI interviewers can effectively build rapport with respondents. Current literature does not yet explore this topic.

\section{Conclusion}

The advent of AI interviewers raises important questions about when and how to leverage these tools. AI interviewers already appear to be able to handle quantitative survey data collection and provide more functionality than current IVR systems while doing so. Early work suggests that they may be able to collect richer qualitative insights than IVR systems as well. 

Our review of current research suggests a few circumstances when AI interviewers may be particularly useful:
\begin{itemize}
    \item Human interviewer availability is limited (e.g., off hours)
    \item Follow-up questions are not pivotal (i.e., limited complexity of probing needed)
    \item Post-processing of respondent data (e.g., audio recordings) is acceptable (to minimize transcription errors)
    \item Emotion detection and expression are not central to the study
    \item Sensitive topics (e.g., money, sexuality) are involved that may trigger social desirability bias in responses to a human interviewer
\end{itemize}

In such scenarios, AI interviewers may be able to provide quantitative and qualitative insights cost-effectively.

\subsection{Future Research}
We are still in the early days of AI interviewer research. Although insights are beginning to emerge about the capabilities of AI interviewers, significantly more work is needed to measure their performance – especially in relation to human interviewers. We note several key areas for further research:

\begin{itemize}
    \item \textbf{Respondent Representativeness:} Most early work to date has been conducted on college/graduate students. It is unclear whether results about response rates and respondent experience will generalize to a general population. For example, an open question remains about whether higher-levels of antipathy towards AI will emerge as this technology becomes widespread. We still do not know which types of respondents tend to prefer AI interviewers, nor which contexts or topics such preferences are most salient for.
    \begin{itemize}
        \item \citet{leybzon2025ai} experiment with a probability-based panel meant to be representative of the US population, but sample sizes are not large enough to draw conclusive inferences across specific subpopulations.
    \end{itemize}
    \item \textbf{Respondent Preferences:} Early signs are encouraging that respondents are willing to speak with AI interviewers. \citet{leybzon2025ai} find most respondents to be at least equally comfortable speaking with AI versus a human, particularly for sensitive topics, though more research is necessary to confirm this hypothesis. Interestingly, \citet{wuttke2024ai} find that respondents may have a lower tolerance for conversational latency when made aware that they were interacting with an AI interviewer; some respondents actively noted the time necessary for responses to be generated.
    \item \textbf{Methodological Variation:} 
    \begin{itemize}
        \item \textbf{Inbound vs Outbound Campaigns:} Most studies to date involve outbound phone call campaigns with limited inbound options. \citet{lang2025telephone} did offer respondents an option to connect to a web-based call on-demand, but further investigation is necessary to see whether the act of opting-into a survey meaningfully changes results. Offering voice-based interview options alongside traditional web form surveys or SMS campaigns has also not yet been fully explored.
        \item \textbf{Interviewer Attributes:} Respondents may have different responses to variations in interviewer behavior (e.g., use of humor, small talk, and affirmation) as well as which accents/voices are used (see \citet{yun2023keeping}, \citet{lilley2025social}). The impact of such changes on respondent experience feedback remains to be seen.
    \end{itemize}
\end{itemize}

 We acknowledge that the speed at which AI technologies have developed has been rapid; it is possible that the capabilities of AI interviewers will progress quickly in the coming months and years. Investigation of this exciting new modality will be crucial to maintain methodological rigor and inform judgments on the many ethical issues that are sure to arise from this technology's widespread usage.

\bibliography{colm2025_conference}
\bibliographystyle{colm2025_conference}
\end{document}